\documentclass[10pt,twocolumn,letterpaper]{article}

\usepackage{iccv}
\usepackage{times}
\usepackage{epsfig}
\usepackage{graphicx}
\usepackage{amsmath}
\usepackage{amssymb}

\usepackage{graphicx}
\usepackage{amsmath}
\usepackage{amssymb}
\usepackage{booktabs}
\usepackage{subcaption}
\usepackage{enumitem}
\usepackage{multirow}
\usepackage{amssymb}%
\usepackage{pifont}%
\newcommand{\cmark}{\ding{51}}%
\newcommand{\xmark}{\ding{55}}%

\usepackage[pagebackref=true,breaklinks=true,letterpaper=true,colorlinks,bookmarks=false]{hyperref}

\usepackage[capitalize]{cleveref}
\crefname{section}{Sec.}{Secs.}
\Crefname{section}{Section}{Sections}
\Crefname{table}{Table}{Tables}
\crefname{table}{Tab.}{Tabs.}

\iccvfinalcopy %

\def\iccvPaperID{3} %

\ificcvfinal\fi

\begin{document}

\title{On the Interplay of Convolutional Padding and Adversarial Robustness}

\author{
Paul Gavrikov$^{1}$\thanks{Funded by the German Ministry for Science, Research and Arts, Baden-Wuerttemberg under Grant 32-7545.20/45/1 (Q-AMeLiA).} \qquad Janis Keuper$^{1,2}$\footnotemark[1]\\
$^{1}$IMLA, Offenburg University, $^{2}$Fraunhofer ITWM\\
{\tt\small \{paul.gavrikov,janis.keuper\}@hs-offenburg.de}
}

\maketitle
\ificcvfinal\thispagestyle{empty}\fi

\begin{abstract} %
It is common practice to apply padding prior to convolution operations to preserve the resolution of feature-maps in Convolutional Neural Networks (CNN). While many alternatives exist, this is often achieved by adding a border of zeros around the inputs. In this work, we show that adversarial attacks often result in perturbation anomalies at the image boundaries, which are the areas where padding is used. Consequently, we aim to provide an analysis of the interplay between padding and adversarial attacks and seek an answer to the question of how different padding modes (or their absence) affect adversarial robustness in various scenarios.
\end{abstract}

\section{Introduction}
\label{sec:intro}
\begin{figure}
    \centering
    \includegraphics[width=\columnwidth]{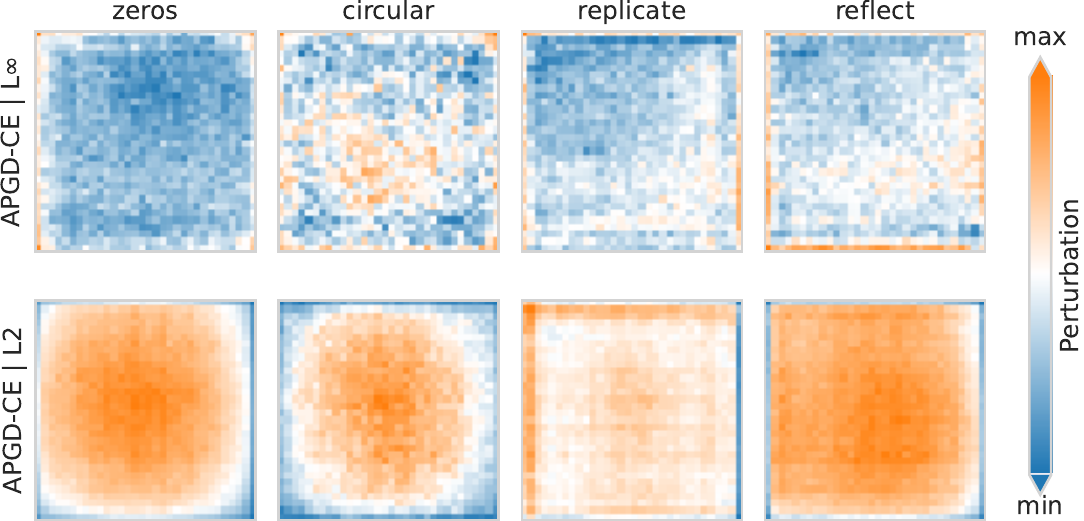}
     \caption{Average perturbations of 1,000 \textit{CIFAR-10} samples where attacks were successful. Adversarial attacks show perturbation anomalies (more or fewer perturbations) at image boundaries that can be attributed to padding. Shown here is a \textit{ResNet-20} trained without adversarial defenses with different padding modes under attacks of high-budget $\ell_\infty$- (top row) and $\ell_2$-bounded (bottom row) \textit{APGD-CE}.}
    \label{fig:padding_anomaly}
\end{figure}

Over the recent years, Convolutional Neural Networks (CNN) \cite{lecun1995convolutional} have become the dominant backbone of most learning-based approaches for computer vision applications \cite{zarandy2015overview}. However, despite their overwhelming success in terms of achieving high test accuracies on various (vision) benchmarks, CNNs also have shown to be very vulnerable against minor changes in the input data distribution, \eg, against adversarial attacks \cite{akhtar2018threat}. This is especially concerning in safety-critical applications such as autonomous driving \cite{cao2019} or medical imaging \cite{Finlayson2019} where human lives are at stake.
Beyond adversarial training \cite{madry2018towards}, which can be considered as the current standard approach towards the training of more robust networks, recent works showed that inherent signal processing flaws related to convolutions are one of several possible sources for the lack of robustness. 
While prior investigations, including down-sampling \cite{julia1,Grabinskilowcut22}, the size of the convolution kernels \cite{durall2020watch, tomen2021spectral} and the choice of the nonlinear activation functions \cite{xiao2019training,weng2018towards}, have shown a significant impact of these network components on the robustness of trained CNN models, there is no systematic analysis of the commonly applied padding schemes in this context.\\
Our contribution is motivated by an initial analysis of spatial attack locations (see fig. \ref{fig:padding_anomaly}), which showed strong anomalies of attack intensities at the boundaries of attacked images. Hence, we investigate the role of padding and its parameters (like type and size) for the robustness of CNNs.\\  
\noindent The key contributions of this paper are:

\begin{itemize}
\item We provide the first in-depth analysis of the impact of the padding-related architectural design choices in the context of adversarial CNN robustness.    
\item Our empirical evaluations on \textit{CIFAR-10} \cite{cifar10} show that the commonly applied {\it same} sized {\it zero} padding does not always result in the best performance, especially in combination with adversarial training. However, standard benchmarks such as \textit{AutoAttack} \cite{croce2020reliable} fail to reflect this.
\item Additionally, we also investigate padding-free architectures, \ie, by using up-scaling or out-painting to increase the input image/feature-map size to compensate for the down-scaling effect of non-padded convolutions. 
\end{itemize}
\begin{figure}
    \centering
    \resizebox{\columnwidth}{!}{
         \begin{tabular}{cccc}
         \includegraphics[width=0.25\columnwidth]{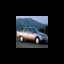} & \includegraphics[width=0.25\columnwidth]{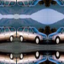} & \includegraphics[width=0.25\columnwidth]{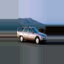} & \includegraphics[width=0.25\columnwidth]{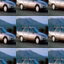} \\
         \textbf{zeros} & \textbf{reflect} & \textbf{replicate} & \textbf{circular} \\
         \end{tabular}
     }
     \caption{Examples of a \textit{CIFAR-10} training sample increased to an image resolution to $64\times64$ px by different padding approaches.}
    \label{fig:padding_modes}
\end{figure}
\section{Related Work}
\label{sec:related_work}
\paragraph{Adversarial robustness.}
Neural networks tend to overfit the training data distribution and fail to generalize beyond. As such the predictions are often highly sensitive to small input perturbations \cite{biggio2013,SzegedyZSBEGF13} that are (almost) imperceptible and semantically meaningless to humans. In some cases, these perturbations can be as small as a single pixel \cite{8601309}. Formally, this phenomenon can be introduced as follows. Given a model $\mathcal{F}$ parameterized by $\theta$, an input sample $x$ with the corresponding class label $y$, and a loss function $\mathcal{L}$, an adversarial attack will attempt to maximize the loss $\mathcal{L}$ by finding an additive perturbation to $x$ in the $\mathcal{B}_{\epsilon}(x)$ ball that is centered at $x$. The $\ell_p$-norm denoted by $\|\cdot\|_p$ is bounded by the radius (budget) $\epsilon$ to restrict perturbations to minor changes.
\begin{equation}
    \begin{gathered}
        \displaystyle\max_{x'\in \mathcal{B}_{\epsilon}(x)}\mathcal{L}\left(\mathcal{F}\left(x'; \theta\right), y\right)\\
        \mathcal{B}_{\epsilon}(x) = \{x':\|x-x'\|_{p}\leq\epsilon\}
    \end{gathered}
\end{equation}
Adversarial attacks can be found in both, white- and black-box settings \cite{liu2017delving,ilyas2018blackbox,Bhagoji2018Oct,andriushchenko2020}. Amongst the most effective attacks are gradient-based white-box attacks that use the model prediction to perturb images in the direction of the highest loss \cite{goodfellow2015explaining,madry2018towards,croce2020reliable,croce2020minimally}. %
Models trained without adversarial defenses can typically not withstand attacks with high $\epsilon$ budgets. An unequivocal solution to overcome this phenomenon is \textit{adversarial training (AT)} \cite{madry2018towards}. Adversarial training trains the model on worst-case perturbations found during training and effectively turns out-of-domain attacks to in-domain samples. As a side-effect, this results in models that classify based on shapes and not texture information which is better aligned with human vision \cite{Gavrikov_2023_CVPRW,geirhos2018imagenettrained}. Further, they are less over-confident \cite{grabinski2022robust} than normally trained ones. 
Unfortunately, adversarial training is also susceptible to overfitting to the attacks employed during its training phase \cite{Wong2020Fast,NEURIPS2020_b8ce4776,Kim_Lee_Lee_2021}. %
Hence, a common choice to assess robustness is \textit{AutoAttack} \cite{croce2020reliable} which compares against multiple attacks \cite{carlini2019evaluating} such as \textit{APGD-CE} \cite{croce2020reliable}, \textit{FAB} \cite{croce2020minimally}, and \textit{Square} \cite{andriushchenko2020}.
\paragraph{Correlation between architecture and robustness.}
\textit{Tang}~\etal~\cite{Tang2021} performed a mass evaluation of different networks against robustness. They concluded that architecture plays an important role in robustness but no universal training recipe exists. For example, they recommend training light-weight architectures with AdamW \cite{loshchilov2018decoupled}, but SGD \cite{Robbins1951ASA} performs significantly better for heavier architectures. Further, they state that CNNs outperform transformers on natural and system noise, but the opposite holds for adversarial robustness.
\textit{Huang}~\etal~\cite{Huang2022} tested the influence of topology, kernel size, activation, normalization, and network size of residual networks on robustness. They find that under the same FLOPs, narrow-deep networks are more robust than shallow-wide networks and derive an optimal ratio. Further, they observe that pre-activation increases robustness, but increasing kernel size does not. Robust architectures can also be an optimization goal in Neural Architecture Search \cite{jung2023neural}.
\textit{Gavrikov}~\etal~\cite{Gavrikov_2022_CVPRcnnfilterdb,Gavrikov_2022_CVPRW} provided an analysis in weight space and showed that convolution filters of adversarially trained CNNs are more diverse in learned patterns and generally occupy more of the network capacity than normally trained counterparts.
\paragraph{Padding.}
Prior to convolution, inputs can be artificially enlarged (padding), \eg, to maintain the resolution before and after convolutions. Padding is controlled by two parameters: the \textit{padding mode} which determines the information in the border, and the \textit{padding size} which determines the size of the border. Popular frameworks like PyTorch provide a multitude of padding modes (see \cref{fig:padding_modes} for examples):
\begin{itemize}%
    \item \textbf{zeros:} the padding area is filled with zeros.
    \item \textbf{reflect:} mirrors the input at the boundary.
    \item \textbf{replicate:} copies the last pixel at the image boundary into the padding area. 
    \item \textbf{circular:} wraps the input around the boundary and continues at the opposite boundary(s).
\end{itemize}
Despite the variety in choice, few recent works have studied the importance of padding in convolution layers, and the majority of image classification networks after \textit{AlexNet} \cite{NIPS2012_c399862d} use \textit{zero} padding. 
In theory, the padding size can be set independently on all image axes, most commonly though, it is kept similar constant and determined by the kernel size $k$: $\lfloor k/2 \rfloor$ (also called \textit{same} padding). For instance, for $k=3$ the padding size for \textit{same} padding is 1.
Multiple works showed that \textit{zero} padding increases translational invariance \cite{NEURIPS2020_c4ede56b,Islam2020How,Kayhan_2020_CVPR} and \textit{circular} padding appears to break it \cite{NEURIPS2020_c4ede56b}.
However, the commonly used \textit{same zero} padding can cause the model to produce artifacts in feature-maps that result in the loss of visual features (blind spots) \cite{alsallakh2021mind}. The authors link this to even input resolutions and show that switching to uneven resolutions improves accuracy. Although all other padding modes except zero padding reduce artifacts in feature-maps, the authors cannot recommend a single best padding mode, as the best choice appears to depend on the specific problem \cite{10.1007/978-3-319-46448-0_5,Schubert2019CircularCN,Vashishth_Sanyal_Nitin_Agrawal_Talukdar_2020}.

To the best of our knowledge, no investigation of the role of padding on robustness has been performed so far. We aim to close this gap in this work.
\begin{figure*}
    \centering
    \begin{subfigure}{\columnwidth}
        \includegraphics[width=\linewidth]{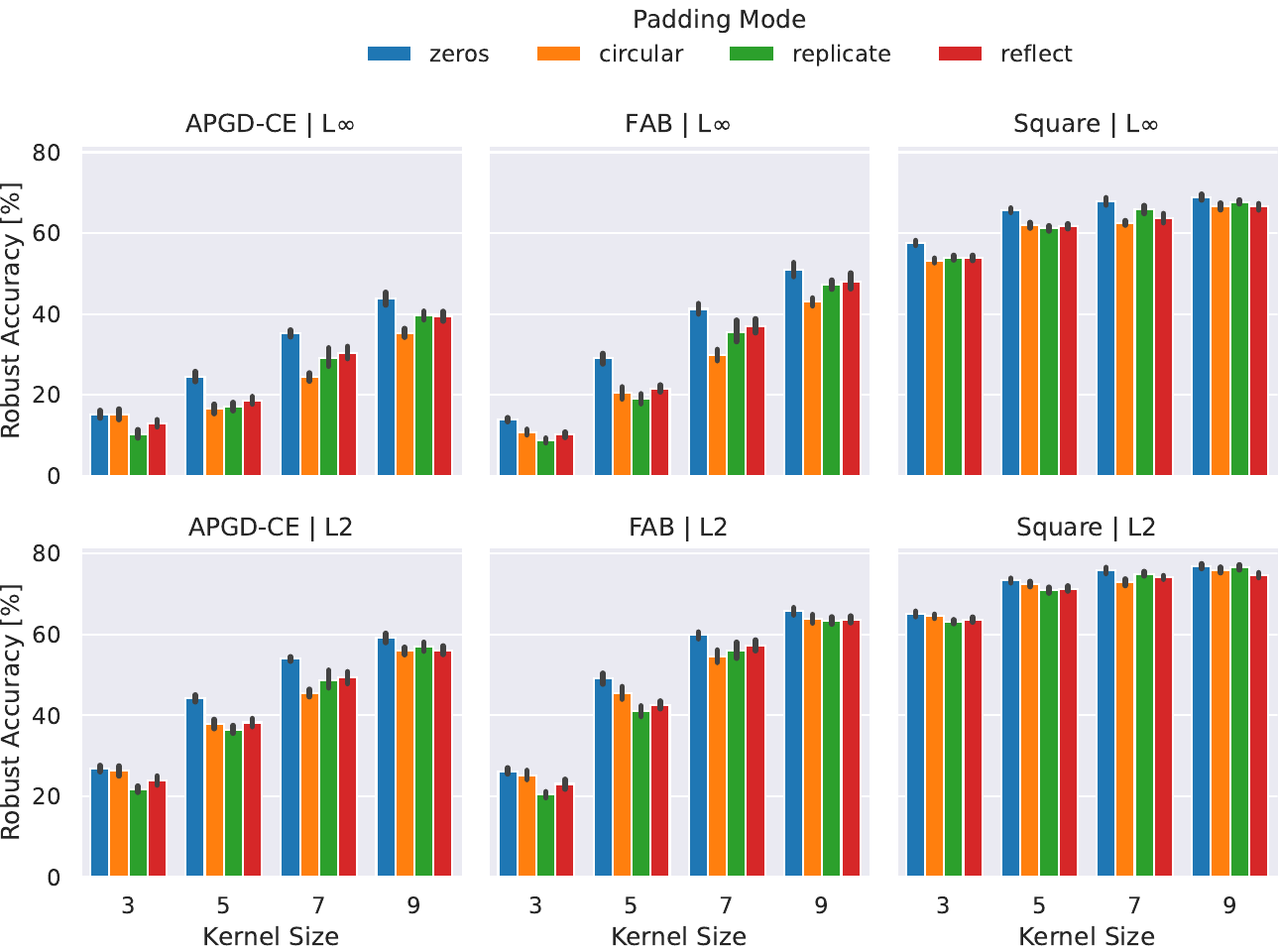}
        \caption{Native robustness (low budget).}
        \label{subfig:kernel_size_vs_robustness_norm}
    \end{subfigure}
    \hfill
    \begin{subfigure}{\columnwidth}
        \includegraphics[width=\linewidth]{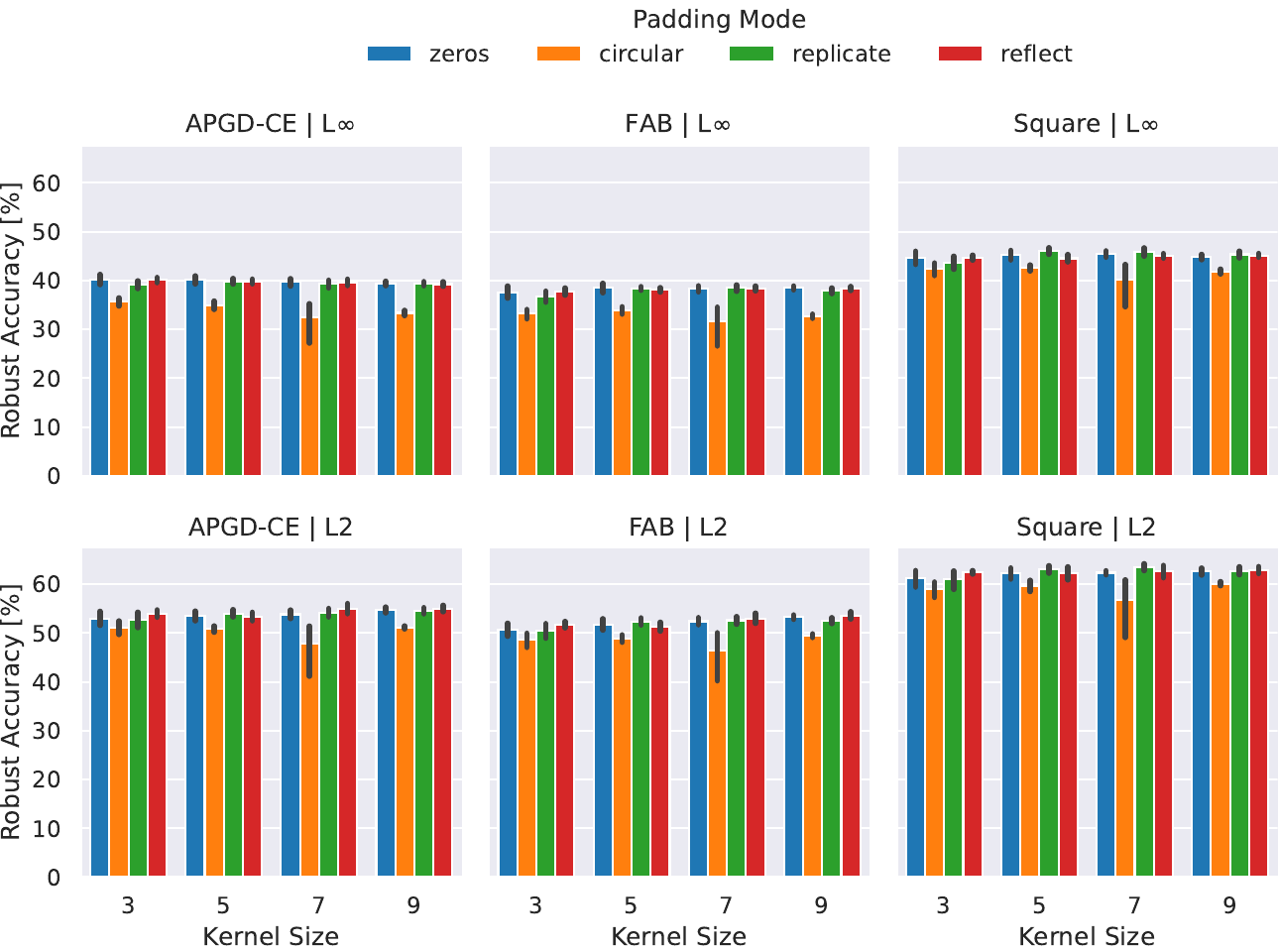}
        \caption{Robustness after adversarial training (high budget).}
        \label{subfig:kernel_size_vs_robustness_at}
    \end{subfigure}    
    \caption{Robust accuracy under different attacks on a \textit{ResNet-20} trained on \textit{CIFAR-10} with different padding modes and kernel sizes. \cref{subfig:kernel_size_vs_robustness_norm} low-budget attacks on normally trained models. \cref{subfig:kernel_size_vs_robustness_at} high-budget attacks on adversarially trained models. Variance computed over 10 models.}
    \label{fig:kernel_size_vs_robustness}
\end{figure*}
\section{Experiments}
\label{sec:experiments}
For our experiments, we train models and switch the padding mode in all convolution layers between \textit{zeros}, \textit{reflect}, \textit{replicate}, and \textit{circular}. Further, we switch the convolution kernel size $k\in\{3,5,7,9\}$ and set the padding size to $\lfloor k/2 \rfloor$ per side, accordingly (\textit{same} padding). The trained models are attacked with \textit{APGD-CE} \cite{croce2020reliable}, \textit{FAB} \cite{croce2020minimally}, and \textit{Square} \cite{andriushchenko2020} using the implementation in \cite{croce2020reliable}. The attacks on models are performed under two different budgets, derived from trends in recent literature. For \textit{low budget} attacks we generate attacks from  $p=2, \epsilon=0.1$ and  $p=\infty, \epsilon=1/255$; and for \textit{high budget} attacks $p=2, \epsilon=0.5$ and $p=\infty, \epsilon=8/255$. Attacks are evaluated on a subset of 1,000 \textit{CIFAR-10} test samples. We measure the clean performance and robust performance under individual attacks and store and analyze perturbed inputs that have successfully fooled the model.
\paragraph{Training Details.}
We train 20-layer deep ResNets optimized for \textit{CIFAR-10} \cite{cifar10} (\textit{ResNet-20}) as introduced in \cite{resnet} models on \textit{CIFAR-10} \cite{cifar10} with the default train/validation splits. Training images are randomly horizontally flipped during training. Test images are not modified. Both splits are normalized by the channel mean and standard deviation. Training is executed with an SGD \cite{Robbins1951ASA} optimizer (with Nesterov momentum \cite{Nesterov1983} of 0.9) for 75 epochs, with an initial learning rate of 0.01 following a cosine annealing schedule \cite{loshchilov2017sgdr}, a weight decay \cite{Krogh1991} of 0.01, a batch size of 256, and cross-entropy loss \cite{good1952} with a label smoothing \cite{Goodfellow-et-al-2016} value of 0.1. For our analyses, we use the model checkpoints at the end of training. For adversarial training experiments, we follow the same parameters except that we train with an FGSM \cite{goodfellow2015explaining} $p=\infty, \epsilon=8/255$ adversary and select the checkpoint with the highest accuracy against a PGD \cite{kurakin2017adversarial} $p=\infty, \epsilon=8/255$ adversary (early stopping) \cite{goodfellow2015explaining} on the test data to prevent \textit{robust overfitting} \cite{Rice2020,Wong2020Fast,NEURIPS2020_b8ce4776}. Although neither our training nor tested architecture is comparable with SOTA approaches (\eg, \cite{kang2021stable,rebuffi2021data,gowal2021improving,Gowal2020,Huang2022}), we believe it to be sufficiently well suited for this analysis. Unless stated otherwise we report results over 10 model runs with different randomness seeds.
\begin{table}
    \centering
    \caption{Clean test accuracy of \textit{ResNet-20} trained on \textit{CIFAR-10} with different padding modes, kernel sizes, and adversarial training. Mean over 10 runs. \textbf{Best}, \underline{second best}. We do not report the second best option for all normally trained runs with $k<9$ as the differences between non-\textit{zeros} padding modes are insignificant.}
    \label{tab:clean_acc}
    \small
    \begin{tabular}{c|c|cccc}
\toprule
      &  &  \multicolumn{4}{c}{\textbf{Clean Test Performance} [\%] ($\uparrow$)} \\
AT & $k$ & zeros &  circular &  replicate &  reflect  \\
\midrule
\multirow{4}{*}{\xmark} & 3 &  \textbf{90.26} &     90.10 &      90.13 &    {90.15} \\
      & 5 &  \textbf{90.14} &     89.66 &      {89.82} &    89.67 \\
      & 7 &  \textbf{89.36} &     88.49 &      {88.52} &    88.47 \\
      & 9 &  \textbf{88.22} &     \underline{87.50} &      87.03 &    87.25 \\
\midrule
\multirow{4}{*}{\cmark}  & 3 &  \underline{71.84} &     69.17 &      70.79 &    \textbf{73.11} \\
      & 5 &  \underline{73.72} &     71.34 &      \textbf{74.02} &    73.08 \\
      & 7 &  \underline{73.86} &     67.33 &      \textbf{73.89} &    73.10 \\
      & 9 &  \underline{73.51} &     71.53 &      72.24 &    \textbf{73.90} \\
\bottomrule
    \end{tabular}
\end{table}
\begin{table}
    \centering
    \caption{\textit{AutoAttack} \cite{croce2020reliable} robust test accuracy of \textit{ResNet-20} trained on \textit{CIFAR-10} with different padding modes, kernel sizes, and adversarial training. Mean over 10 runs. \textbf{Best}, \underline{second best}.}
    \label{tab:aa_acc}
    \small
    \begin{tabular}{c|c|cccc}
\toprule
      &  &  \multicolumn{4}{c}{\textbf{AutoAttack Performance} [\%] ($\uparrow$)} \\
AT & $k$ & zeros &  circular &  replicate &  reflect  \\
\midrule
\multirow{4}{*}{\xmark} & 3           & \textbf{8.52}  & 4.69     & 4.90      & \underline{5.79}    \\
                        & 5           & \textbf{17.69} & 10.44    & 11.12     & \underline{12.33}   \\
                        & 7           & \textbf{29.06} & 17.86    & \underline{24.55}     & 24.35   \\
                        & 9           & \textbf{39.18} & 30.52    & \underline{36.39}     & 34.81   \\ \midrule
\multirow{4}{*}{\cmark} & 3           & \textbf{36.88} & 32.09    & 35.91     & \underline{36.82}   \\
                        & 5           & \textbf{37.48} & 32.34    & \underline{37.30}     & 37.12   \\
                        & 7           & \textbf{37.42} & 30.16    & 37.08     & \underline{37.26}   \\
                        & 9           & \textbf{37.49} & 31.09    & 36.89     & \underline{37.25}   \\
\bottomrule
\end{tabular}
\end{table}
\subsection{Clean Performance}
On clean data and without adversarial training, we see that \textit{zero} padding outperforms all other padding modes in clean accuracy (\cref{tab:clean_acc}). However, under adversarial training \textit{replicate} or \textit{reflect} outperform \textit{zero} padding, depending on the kernel size. While the difference is not very high with larger kernels, at the common kernel size $k=3$ it amounts to 1.27\%. By a large margin, \textit{circular} padding performs the worst in combination with adversarial training at all kernel sizes. We also observe that the accuracy falls off with increasing kernel size without adversarial training, while there is hardly a correlation on adversarially-trained models. However, it is worth noting that $k>3$ models appear to reach higher accuracy under adversarial training.
\subsection{Robust Performance} Next, we compare the robust accuracy under different attacks while testing different padding approaches and kernel sizes (\cref{fig:kernel_size_vs_robustness}).
\paragraph{Native Robustness.}
First, we analyze the performance of models trained without adversarial defenses such as adversarial training. %
For low budgets, we observe that \textit{zeros} outperforms all other padding modes in most settings by a large margin (\cref{subfig:kernel_size_vs_robustness_norm}). %
Generally, we see a very similar trend for padding modes independent of the individual attack or norm, of course, except for the actual robust performance (\textit{FAB} is the strongest attack, and \textit{Square} is the weakest). Most notably, we observe that increasing $k$ seems to diminish the gap between \textit{zero} padding and other modes. Interestingly, contrary to the observations of \cite{Huang2022} we find that increasing kernel size also significantly improves robustness in this setting. For example, we see an improvement by 28.64\% between $k=3$ and $k=9$ kernels against $\ell_\infty$-bounded \textit{APGD-CE} adversaries. As expected, when switching to high-budget attacks (not pictured for brevity), all tested padding mode/kernel size combinations collapse below random performance and mostly even to near 0 performance. For $k=3$, \textit{circular} seems to be the only interesting outlier, as it gains a few percent in robust accuracy, yet, still underperforms a random baseline.
\paragraph{Adversarial Training.}
Additionally to native robustness, we aim to understand whether differences arise in models trained with adversarial training (\cref{subfig:kernel_size_vs_robustness_at}). Contrary to previous results, observing patterns in this setting becomes less straightforward. Averaging over all attacks, we again see that robustness improves with increasing kernel size, albeit at an almost insignificant rate for $k\geq5$. Additionally, all padding modes seem to perform reasonably well but depending on the attack, norm, or kernel size individual modes perform better than others. The only exception to this is \textit{circular} padding. Although it performed reasonably well regarding native robustness, in combination with adversarial training it is always by far the worst choice. 
Regarding the other padding modes, we make the following observations: If we average the robust performance over all attacks and norms we see the same patterns as for clean accuracy, \textit{reflect} is the best choice for $k=3,9$, \textit{replicate} performs best for $k=5,7$. However, we also see large fluctuations. For example, for $k=5$ \textit{reflect} becomes the worst choice after \textit{circular} with \textit{replicate} being the best, on $k=7$ the difference between \textit{reflect} and \textit{replicate} is marginal as both perform almost equally well and, finally, at $k=9$ we see a switch and \textit{replicate} becomes the worst choice after \textit{circular} while \textit{reflect} becomes the best choice. Interestingly, \textbf{\textit{zero} padding is never the best option on average over all attacks}. Yet, if we only analyze $\ell_\infty$-bounded \textit{APGD-CE} attacks, it outperforms all other padding modes by a non-negligible margin. This is a concern when evaluating with \textit{AutoAttack} \cite{croce2020reliable} which starts with \textit{APGD-CE} attacks, and proceeds with other attacks only on samples that could not be successfully attacked. As such it is not surprising that in terms of \textit{AutoAttack} accuracy, \textit{zero} padding remains the best-performing method \cref{tab:aa_acc}. However, \textit{replicate/reflect} padding modes are usually only marginally worse and come at the benefit of improved clean accuracy.
\begin{figure*}
    \centering
    \begin{subfigure}{\columnwidth}
        \includegraphics[width=\linewidth]{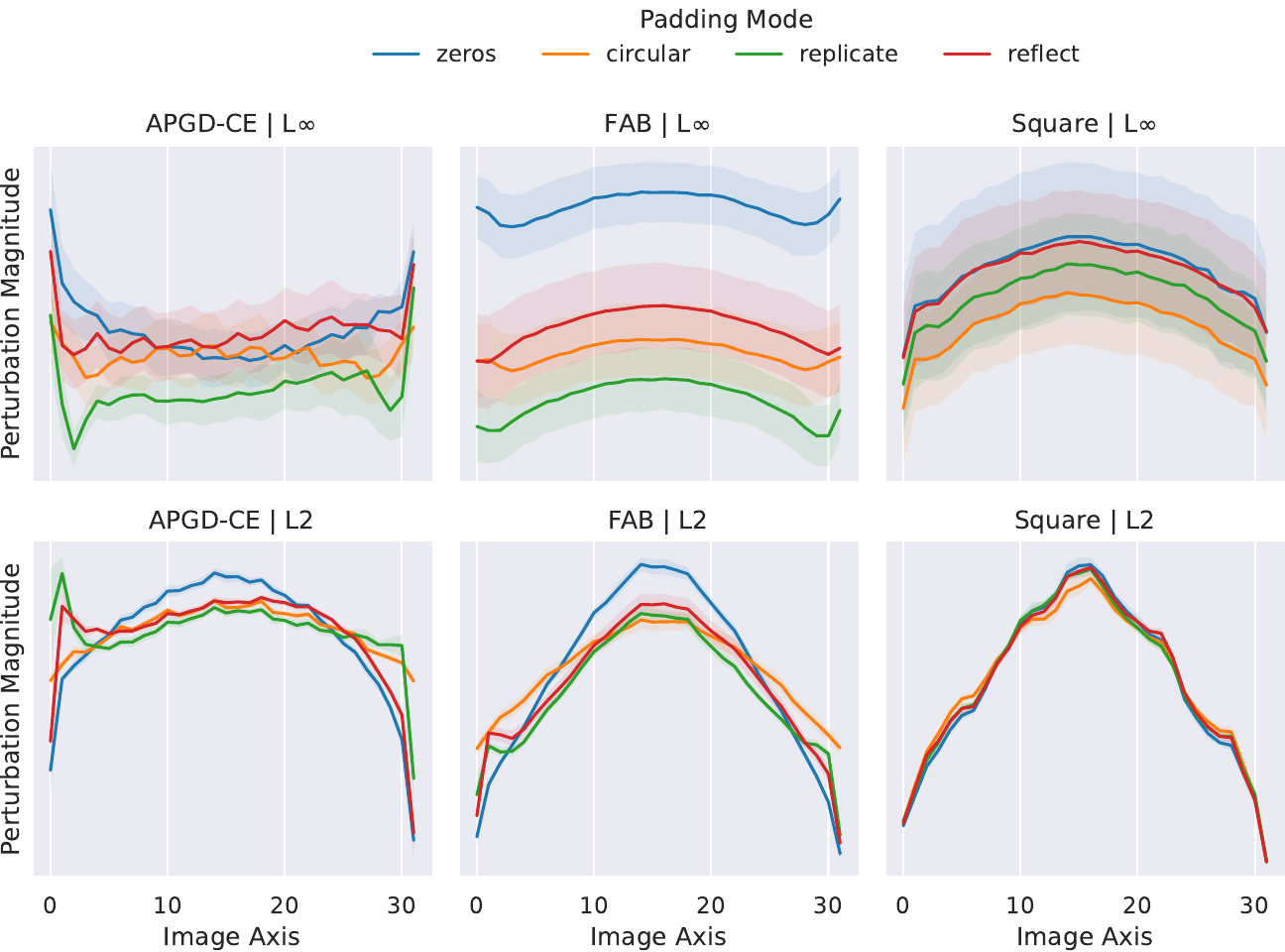}
        \caption{Normal training (high budget attack).}
        \label{subfig:attack_proj_norm}
    \end{subfigure}
    \hfill
    \begin{subfigure}{\columnwidth}
        \includegraphics[width=\linewidth]{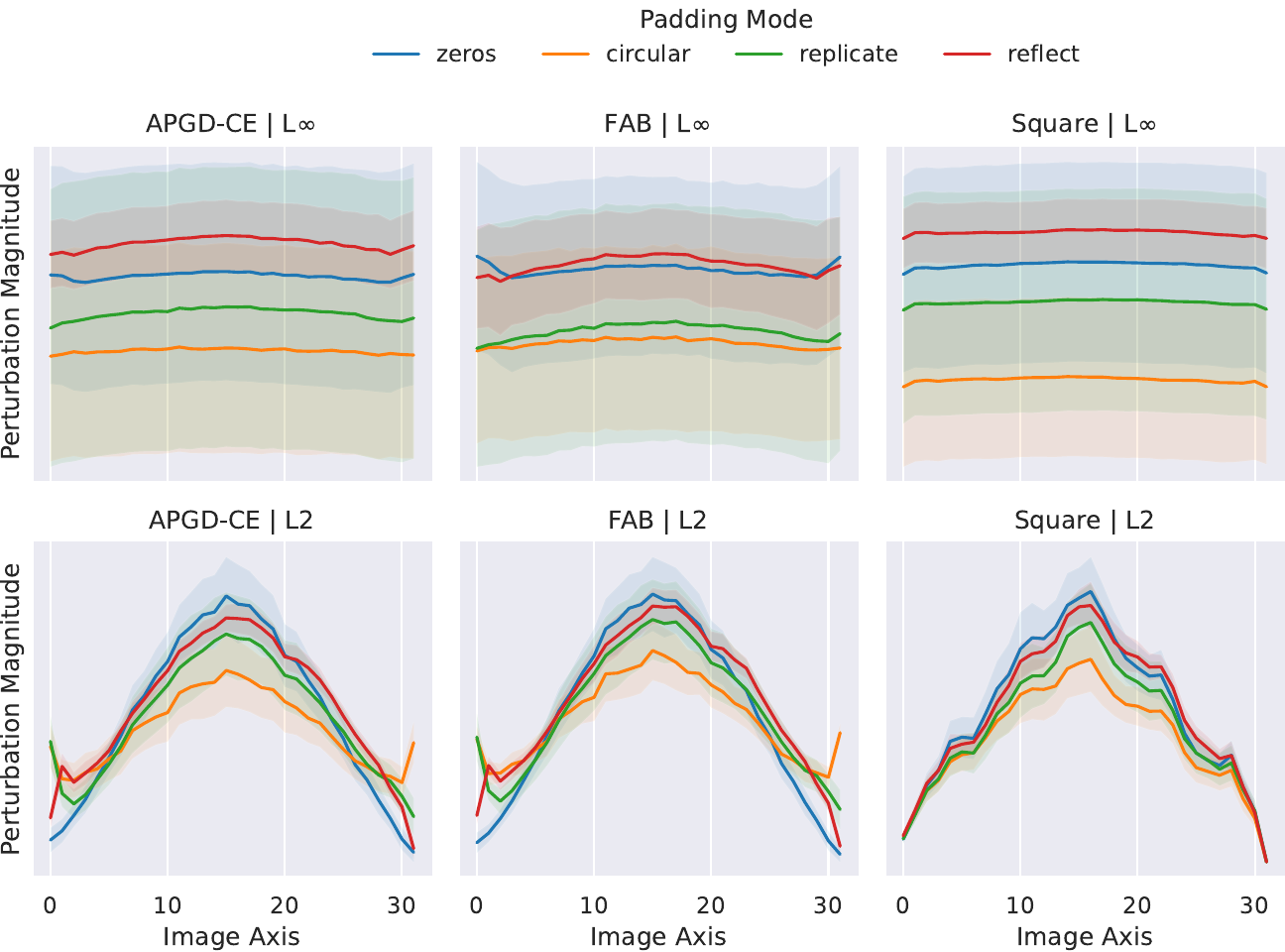}
        \caption{Adversarial training (high budget attack).}
        \label{subfig:attack_proj_at}
    \end{subfigure}
    \caption{Distribution of perturbation magnitudes of successful attacks along the image X-axis under different padding modes, and different attacks on $k=3$ models. \cref{subfig:attack_proj_norm} shows high-budget attacks on normally trained models and \cref{subfig:attack_proj_at} high-budget attacks on adversarially trained models. Variance computed over 10 models.}
    \label{fig:attack_proj}
\end{figure*}
\subsection{Padding Anomalies} 
Following our robustness performance analysis, we now aim to understand where perturbations are primarily located and how different padding modes influence their distribution on $k=3$ models. To this end, we average the absolute differences between successful adversarial examples (i.~e. those that flip the predicted label) and the corresponding clean samples (\cref{fig:padding_anomaly}). We separate between normally-trained (native) models and adversarially-trained models, but this time evaluate both under high budgets to increase perturbations aiming to improve visibility.
For a better comparison between tested dimensions, we project perturbation magnitudes to the image X-axis  (\cref{fig:attack_proj}). Generally, though, we see relatively similar observations on the Y-axis, except that perturbations are less symmetrically distributed and increase towards the lower image edge.
\paragraph{Native Models.}
We see significant differences (\cref{subfig:attack_proj_norm}) between $\ell_2$-perturbations which are primarily located in the center of the image and $\ell_\infty$-perturbations which appear to be distributed more uniformly across the image but contain anomalies in outermost pixels or their close neighbors. Noticeably, these are areas where the receptive field intersects with the padded area. For $\ell_\infty$-bounded \textit{APGD-CE}, and to a lesser extent also \textit{FAB}, we see increases in perturbations at boundaries, while \textit{Square} shows strong decreases in perturbations at boundaries. Generally, we see similar distribution shapes independent of padding mode, except for $\ell_2$-bounded \textit{APGD-CE} where \textit{reflect/replicate} show increased perturbations at the boundaries. 
However, we see a different area under the curve (AUC) for padding modes and \textit{zero} shows the largest AUC.
\paragraph{Adversarial Training.}
Under adversarial training (\cref{subfig:attack_proj_at}) perturbation distributions of $\ell_\infty$-bounded attackers level out but show a significantly increased variance between runs. Anomalies at the boundaries vanish for the largest part but are still slightly noticeable on all padding modes except \textit{circular}. \textit{Reflect} now has the largest AUC. For $\ell_2$-bound adversaries we see relatively similar distributions, even in AUC, except for \textit{circular} which has the lowest AUC overall. Boundary anomalies are again visible for \textit{replicate/reflect}, but now also for \textit{circular}.
\subsection{The Computational Overhead of Padding}
We have seen that \textit{zero} padding may not always be the best choice, depending on the training type and the attack used. However, it has one significant advantage over the other padding modes used as it is entirely independent of the image content. Theoretically, this would mean that \textit{zero} padding would be the computationally cheapest option. To understand if this also holds on real hardware that often optimizes processes we benchmark the required time for padding. In addition to just padding we also measure the time for a 2D convolution (32 filters with $3\times 3$ kernel size and 1 px padding on all sides) operation on single input ($32\times 32$ resolution) for a more realistic scenario. Both operations are evaluated on GPU. We run the experiments on an NVIDIA A100-SXM4-40GB GPU with CUDA 11.3, cuDNN 8302, using PyTorch 1.12.1. All measurements are reported over 10,000 independent trials.\\
\begin{table}
    \centering
    \caption{Benchmark of the average time for padding or padded 2D convolution operation under different padding modes. Bold marks \textbf{best}.}
    \label{tab:timings}
    \small
    \begin{tabular}{lrrrr}
    \toprule
    & \multicolumn{4}{c}{\textbf{Average time for operation} [$\mu s$] ($\downarrow$)}\\
    \textbf{Operation} & zeros &  reflect &  replicate &  circular  \\
    \midrule
    {Only padding} & 21.87 & 12.55 & \textbf{10.10} & 56.96 \\
    \midrule
    {2D Convolution} & \textbf{55.65} & 76.86 & 74.04 & 132.60\\
    \bottomrule
    \end{tabular}
\end{table}
The results in \cref{tab:timings} show an interesting trend. For the padding operation alone, \textit{zeros} doubles the required time in comparison to \textit{reflect} and \textit{replicate}, but requires only a third of the time of \textit{circular}. However, in terms of the total budget for the convolution \textit{zeros} yields the fastest forward pass. \textit{Reflect} and \textit{replicate} are approximately 35\% slower, and \textit{circular} introduces an overhead of approximately 137\% - presumably, due to optimized implementations. Our results show that the superiority of \textit{replicate}, \textit{reflect} in some settings comes at the cost of slower test and thus also train time.
\subsection{Effect on Model Decisions}
\begin{figure}
    \centering
    \begin{subfigure}{\linewidth}
        \includegraphics[width=\linewidth]{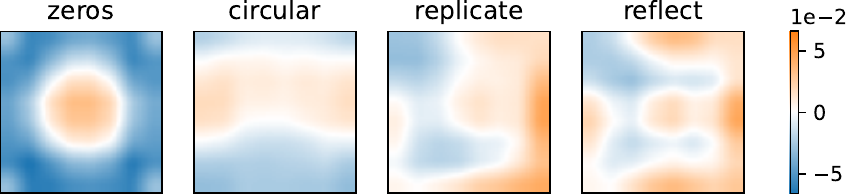}
        \caption{Normal training (low budget attack).}
        \label{subfig:xai_norm_low}
    \end{subfigure}\\
    \begin{subfigure}{\linewidth}
        \includegraphics[width=\linewidth]{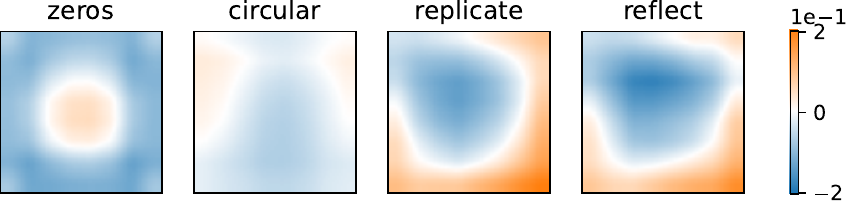}
        \caption{Normal training (high budget attack).}
        \label{subfig:xai_norm_high}
    \end{subfigure}\\
    \begin{subfigure}{\linewidth}
        \includegraphics[width=\linewidth]{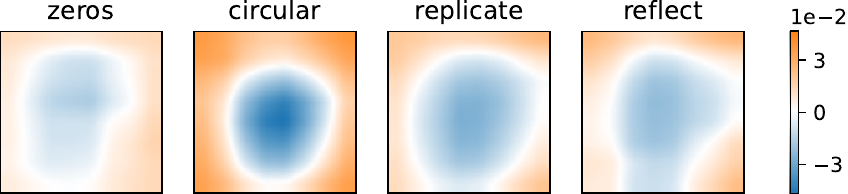}
        \caption{Adversarial training (high budget attack).}
        \label{subfig:xai_at_high}
    \end{subfigure}
    \caption{Average shifts in \textit{LayerCAM} \cite{jiang2021layercam}  explanations following successful $\ell_\infty$-bounded \textit{APGD-CE} attacks under different padding modes. Mean computed over 10 models. Orange areas indicate areas where explanations move under adversarial attacks, while blue areas indicate areas of explanations have been removed.}
    \label{fig:causlity}
\end{figure}

Our previous results showed that the choice of padding method and size can have significant effects on both, the model robustness (\cref{tab:clean_acc} and \cref{fig:kernel_size_vs_robustness}) and the placement of the perturbations (\cref{fig:padding_anomaly}).   
To further analyze whether the perturbation anomalies at image edges are affecting the model decision intrinsics, and thus the robustness, we analyze visual explanations via \textit{LayerCAM} \cite{jiang2021layercam} as implemented in \cite{jacobgilpytorchcam}. Explanations are computed for the feature-maps after the last residual block with respect to the predicted label. We compute the differences in explanations between an adversarial example and its clean counterpart whenever the attack was successful and visualize the mean over all samples of 10 differently seeded models (\cref{fig:causlity}). For this analysis, we limit ourselves to $\ell_\infty$-bounded high-budget \textit{APGD-CE} attacks on $k=3$ models.
There is a clear shift in explanations between normally trained and adversarially-trained models.
For normally trained models under low-budget attacks (\cref{subfig:xai_norm_low}) we observe a difference in the explanation shift depending on the padding mode. With \textit{zero} padding, explanations clearly shift toward the center indicating that the attack actually concentrates on the image foreground (as \textit{CIFAR-10} is well-centered). Under \textit{circular} padding the shift area seems to stretch horizontally and attacks start to aim at image boundaries. \textit{Replicate} and \textit{reflect} padding modes show almost similar behavior: while there is some shift towards the center, there are increasingly more attacks on the image boundary (except for the top left corner) than with previous modes. As we increase the attack budget (\cref{subfig:xai_norm_high}) we observe the same behavior for all padding modes except \textit{zero} padding. Attacks move entirely to the image boundaries padding showing that the adversary attacks everything \textbf{except} the image foreground. The shift intensity is smaller for \textit{circular} than for \textit{replicate/reflect}. For \textit{zero} padding we observe hardly any differences to the low-budget setting.
For adversarially-trained models (\cref{subfig:xai_at_high}) we see quite the opposite: attacks for all padding modes are shifted toward the image edges, with no clear difference between padding modes except for the increased intensity of shifts in \textit{circular} models.
Although not pictured here for brevity, we observe similar trends for $\ell_2$-bounded attacks.
\subsection{Is No Padding a Better Alternative?}
\label{sec:paddingfree}
\begin{figure}
    \centering
    \resizebox{\columnwidth}{!}{
         \begin{tabular}{@{}lcccc@{}}
         \rotatebox{90}{\textbf{\phantom{blau}Clean}} &
         \includegraphics[width=0.25\columnwidth]{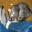} & \includegraphics[width=0.25\columnwidth]{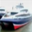} & \includegraphics[width=0.25\columnwidth]{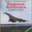} &
         \includegraphics[width=0.25\columnwidth]{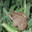} 
         \\
         \rotatebox{90}{\textbf{\phantom{b}Outpainted}} &
         \includegraphics[width=0.25\columnwidth]{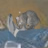} & \includegraphics[width=0.25\columnwidth]{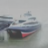} & \includegraphics[width=0.25\columnwidth]{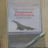} &
         \includegraphics[width=0.25\columnwidth]{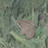} 
         \end{tabular}
     }
     \caption{Examples of \textit{CIFAR-10} validation samples (top row) increased to an image resolution to $48\times48$ px by \textit{MAT} \cite{li2022mat} (bottom row).}
    \label{fig:out_examples}
\end{figure}

In this next section, we aim to understand the role of padding per se on robustness by entirely removing padding from all convolution layers, and without padding, models should not develop boundary anomalies.
Without any other changes, this would result in smaller feature-map (representation) resolutions, and, residual networks would experience a size mismatch between input signals propagated between skip connections and the actual convolution paths. To solve the size mismatch after skip connections we zero-pad the processed signal before the summation instead of directly summing the residual and processed signal. Note that this is different from zero-padded convolutions as it does not result in zero-padded inputs to the next operators. Regarding the smaller feature-maps, we compare $k=3$ padding-free networks in 3 settings: 
\begin{itemize}%
    \item \textbf{Unmodified (None):} We use the same $32\times32$ px \textit{CIFAR-10} samples as before. For \textit{ResNet-20} this results in $4\times4$ representations at the deepest layers instead of $8\times8$. 
    \item \textbf{Upscaling (None + Up):} We upscale \textit{CIFAR-10} samples to $48\times48$ px via bilinear interpolation \cite{Hsieh1978} to match the representations of padded networks.
    \item \textbf{Outpainting (None + Out):} To increase the resolution in a meaningful manner we experiment with image outpainting to $48\times48$ px via \textit{MAT} \cite{li2022mat} pre-trained on the \textit{Places} dataset \cite{zhou2018} containing scenic images that should not interfere with the original label (\cref{fig:out_examples}). We outpaint both, train and test samples.
\end{itemize}
Without adversarial training, we see a drop in clean and robust accuracy if we simply disable padding without countermeasures (\cref{tab:padding_free}). With both, upscaling or outpainting, we are able to mitigate the drop in clean accuracy, yet provoke a complete failure in robust accuracy except for \textit{Square} attacks but still at a decreased robustness. Although it is worth noting that $\ell_\infty$-bounded attacks can attack more area for the same budget when increasing the image resolution as we do in upscaling or outpainting, the delta is too large to be caused by only an increased attack area.
Under adversarial training, we see similar trends, but this time we also measure a clear difference between upscaling and outpainting. Surprisingly, outpainting seems even further to reduce robustness. However, we also notice that outpainting seems to affect non-masked areas (the original sample) and tampers with image statistics.
In all cases, disabling padding results in worse performance and is therefore not recommended.
\begin{table}[h]
    \centering
    \caption{Performance comparison of padding-free $k=3$ models against zero padding. Mean over 3 runs.}
    \label{tab:padding_free}
    \resizebox{\columnwidth}{!}{
    \begin{tabular}{@{}l|c|c|ccc|ccc@{}}
    \toprule
          &  & \multicolumn{7}{c}{\textbf{Test Accuracy} [\%] ($\uparrow$)}\\
          &  &  &\multicolumn{3}{c|}{$\ell_2$} & \multicolumn{3}{c}{$\ell_\infty$} \\
        \textbf{Padding} & \textbf{AT} & \textbf{Clean} &  APGD-CE &   FAB & Square & APGD-CE &   FAB & Square \\
        \midrule
        Zeros & \multirow{4}{*}{\xmark} & \textbf{90.26} &\textbf{ 26.87} & \textbf{26.24} &  \textbf{65.10} &   \textbf{15.23} & \textbf{13.87} &  \textbf{57.55} \\
        None &  &  87.19 &16.90 & 19.23 &  56.40 &    9.80 & 13.47 &  49.47 \\
        None + Up &  &  89.69  &3.03 &  1.03 &  50.70 &    0.20 &  0.00 &  29.97 \\
        None + Out &  &  89.69  &3.03 &  1.03 &  50.70 &    0.20 &  0.00 &  29.97 \\
        \midrule
        Zeros & \multirow{4}{*}{\cmark} & \textbf{71.84} &\textbf{52.93} & \textbf{50.72} &  \textbf{61.17} &   \textbf{40.22} & \textbf{37.62} &  \textbf{44.59} \\
        None &  & 67.40 &48.80 & 46.73 &  56.87 &   38.60 & 35.93 &  41.50 \\
        None + Up & & 47.80  &40.33 & 38.47 &  45.60 &   31.50 & 28.00 &  31.03 \\
        None + Out & &  40.62 &32.63 & 30.07 &  38.43 &   29.00 & 24.80 &  26.27 \\
    \bottomrule
    \end{tabular}
    }
\end{table}
\section{Conclusion \& Discussion}
\label{sec:conclusion}
We have evaluated the adversarial robustness of \textit{CIFAR-10} models under different padding modes, kernel sizes, attacks, and training modes.
Based on our results, we can provide the following recommendations:\\
For settings targeting native robustness, we advise using \textit{zero} padding and increasing the kernel/padding size, if possible. We hypothesize that larger kernels are more likely to cause robustness than larger padding sizes.
Switching to other padding modes clearly deteriorates the performance and starts shifting attacks toward image boundaries. \\
When using adversarial training, it can be fruitful to experiment with the \textit{reflect} and \textit{replicate} padding modes. However, commonly models are benchmarked with \textit{AutoAttack} \cite{croce2020reliable} which starts with \textit{APGD-CE} attacks where \textit{zero} padding performs best, and proceeds with other attacks only on samples that could not be successfully attacked. Since \textit{APGD-CE} is usually a very effective attack, the difference measured based on the remaining subset may not be significant to show clear differences between padding modes. However, at negligible impairments of robustness performance, we saw an improved clean accuracy of non-\textit{zero} padding modes. In addition, we suggest reporting adversarial robustness for multiple attacks separately when studying padding.\\
Lastly, in all cases, we can recommend the usage of padding as padding-free architectures performed significantly worse in all investigated scenarios.\\

\noindent \textbf{Limitations.} For completeness, we want to emphasize that we only experimented on \textit{CIFAR-10}. As with many ``toy-datasets'', objects in question are usually perfectly centered in the images. Unfortunately, this applies to most common benchmarking datasets such as \textit{MNIST} \cite{mnist}, \textit{SVHN} \cite{SVHN}, \textit{CIFAR-100} \cite{cifar10}, \textit{ImageNet} \cite{imagenet} etc. As such, we are not confident that we would obtain significantly different results on these datasets. Repeating our experiments on less curated datasets may, however, result in different observations. Further, we only experimented with ResNet-20. While we believe that the results are representative of the popular residual networks in general, we cannot guarantee that our results scale to entirely different architectures.
We aim to bridge these gaps in future work.

{\small
\bibliographystyle{ieeetr_fullname}
\bibliography{main}
}

\end{document}